\documentclass[12pt]{article}

\usepackage{arxiv}

\usepackage[utf8]{inputenc} 
\usepackage[T1]{fontenc}    
\usepackage{hyperref}       
\usepackage{url}            
\usepackage{booktabs}       
\usepackage{amsfonts}       
\usepackage{nicefrac}       
\usepackage{microtype}      
\usepackage{lipsum}
\usepackage{graphicx}
\RequirePackage{amsmath}
\usepackage{tabularx}
\usepackage{algorithm}
\usepackage[noend]{algpseudocode}
\usepackage{graphicx,subcaption}

\newcommand{\vect}[1]{\boldsymbol{#1}}
\newcommand\VRule[1][\arrayrulewidth]{\vrule width#1}

\title{Generative Learning for Slow Manifolds and Bifurcation Diagrams}

\author{ 
Ellis R. Crabtree \\
  Dept. of Chemical and Biomolecular Engineering\\
  Johns Hopkins University\\
  Baltimore, MD, USA\\
  \texttt{ecrabtr2@jhu.edu} \\
  \And
Dimitris G. Giovanis\\
  Dept. of Civil and Systems Engineering\\
  Johns Hopkins University\\
  Baltimore, MD, USA \\
  \texttt{dgiovan1@jhu.edu} \\
   \And
 Nikolaos Evangelou \\
  Dept. of Chemical and Biomolecular Engineering\\
  Johns Hopkins University\\
  Baltimore, MD, USA\\
  \texttt{nevange2@jhu.edu} \\
  \And
Juan M. Bello-Rivas \thanks{Current address: Microsoft Azure Quantum} \\
  Dept. of Chemical and Biomolecular Engineering\\
  Johns Hopkins University\\
  Baltimore, MD, USA\\
  \texttt{jmbr@superadditive.com} \\
   \AND
 Ioannis G. Kevrekidis\thanks{Corresponding author} \\
  Dept. of Chemical and Biomolecular Engineering\\
  Dept. of Applied Mathematics and Statistics \\
  Johns Hopkins University\\
  Baltimore, MD, USA\\
  \texttt{yannisk@jhu.edu}
}

\begin{document}
\maketitle
\begin{abstract}

In dynamical systems characterized by separation of time scales, the approximation of so called ``slow manifolds'', on which the long term dynamics lie, is a useful step for model reduction.
Initializing on such slow manifolds is a useful step in modeling, since it circumvents fast transients, and is crucial in multiscale algorithms (like the equation-free approach) alternating between fine scale (fast) and coarser scale (slow) simulations. 
In a similar spirit, when one studies the infinite time dynamics of systems depending on parameters, the system attractors (e.g., its steady states) lie on bifurcation diagrams (curves for one-parameter continuation, and more generally, on manifolds in state $\times$ parameter space.  Sampling these manifolds gives us representative attractors (here, steady states of ODEs or PDEs) at different parameter values. 
Algorithms for the systematic construction of these manifolds (slow manifolds, bifurcation diagrams) are required parts of the ``traditional'' numerical nonlinear dynamics toolkit.

In more recent years, as the field of Machine Learning develops, conditional score-based generative models (cSGMs) have been demonstrated to exhibit remarkable capabilities in generating plausible data from target distributions that are conditioned on some given label. 
It is tempting to exploit such generative models to produce samples of data distributions (points on a slow manifold, steady states on a bifurcation surface) conditioned on (consistent with) some quantity of interest (QoI, observable).
In this work, we present a framework for using cSGMs to quickly (a) initialize on a low-dimensional (reduced-order) slow manifold of a multi-time-scale system consistent with desired value(s) of a QoI (a ``label'') on the manifold,  and (b) approximate steady states in a bifurcation diagram consistent with a (new, out-of-sample) parameter value.
This conditional sampling can help uncover the geometry of the reduced slow-manifold and/or approximately ``fill in'' missing segments of steady states in a bifurcation diagram. 
The quantity of interest, which determines how the sampling is conditioned, is either known \emph{a priori} or identified using manifold learning-based dimensionality reduction techniques applied to the training data.
\end{abstract}


\section{Introduction}
\label{sec:intro}
\noindent
In the study of dynamical systems, a long standing goal is the efficient sampling and exploration of long term behavior over a usually long (or coarse) timescale of interest. For high-dimensional dynamical systems, the coarse behavior is frequently described by a slow, reduced-order manifold that is parameterized by coarse descriptors of the system, and this manifold contains the equilibrium measures and steady states of the system. In practice, simulations of said systems are quickly attracted to its manifold and once the manifold is reached, the simulation evolves on it. However, the best models available are often at a fine scale, and the simulations may take extended amounts of time to reach the manifold, making the long term behavior, steady states, and the manifold in general, difficult to observe. Additionally, at certain points the manifold may experience bifurcations, resulting in the presence of new steady states for the system. These dynamical systems are multiscale in nature, having fine-scale behaviors that are more practical to model, and interesting behavior at long time scales that are less practical for the established fine-scale models to uncover. To compensate for this gap between the time scale of the available model and the time scale of interest, the development of a reduced order model becomes necessary. 

In theory, one should always be able to derive a reduced order dynamical model, provided that a reduced order manifold exists; 
however, this manifold may be impossible to explicitly derive in practice due to inaccurate or unavailable closures. For systems that experience these impediments, approaches to identify reduced-order manifolds have been developed in first principles modeling \cite{ILDM, CSP, init_slow_man1}, and more recently, data driven methods have also been developed and explored, including Diffusion Maps \cite{COIFMANdmaps, lafon2004diffusion}, ISOMAP \cite{isomap1}, Autoencoders \cite{autoencoder}, Local Linear Embedding \cite{locallinearemb}, novel machine learning based methods \cite{Wan_Sapsis_2018, CHEN2021110666, patsatzis2023slow}, and more. These techniques can identify the manifold, and once the manifold is identified, the most pertinent step to take is the accurate initialization {\em on the manifold} of the system, sidestepping the equilibration times of the fine-scale model. In other words, the tasks at hand are the quick generation of distributions of data that are consistent with the steady states and equilibrium measures of the system. These tasks amount to two distinct objectives: (a) initialization on the slow manifold in the form of a distribution that is consistent with some observable, or (b) initialization on the manifold in the form of distributions of steady states that are consistent with parameters or measurements of the system.

In machine learning, generative models such as score-based generative models (SGMs) \cite{songdiffmodels}, also known as diffusion models, and generative adversarial networks (GANs) \cite{goodfellow2014} are able to generate high-dimensional data instances from latent distributions. Furthermore, conditional variations of these models (cSGMs and cGANs) \cite{songdiffmodels, mirza2014} can generate conditional distributions of high-dimensional data that are conditioned on some coarse variable of the system. The development of these models has revolutionized tasks such as image and video generation and a plethora of other generative tasks. Particularly, conditional diffusion models (and other generative models) have been used to produce data-driven solutions for forward and inverse problems in systems described by partial differential equations \cite{jacobsen2023cocogen, kadeethum2021framework}, for uncertainty quantification and sampling of chaotic dynamical systems \cite{finzi2023userdefined}, and interpolation of large-dimensional data such as videos and seismic data \cite{jain2024video, doi:10.1190/geo2023-0182.1}. The conditional versions of these models perfectly serve as a solution to the task of quickly initializing on manifolds. Provided a label that parametrizes the manifold, a trained generative model can quickly produce a distribution that is consistent with the prescribed value of the label. The label provided can offer distinct benefits: when the labels are specific observations of the system, the produced output is a measure of the system on the slow manifold and when the labels consist of parameters or measurements of the system, the resulting output is a conditional distribution of the steady states of that system (effectively a bifurcation diagram for the labeled parameters or measurements).
Our previous work also shows that generative models exhibit considerable parallels with long standing techniques from traditional physics-based enhanced sampling, and that machine learning and traditional methods can be used in tandem to initialize multiscale simulations and computations for expedited exploration of phase space \cite{GANs_closures, crabtree2024micro}. Furthermore, capitalizing on the data-driven identification of manifolds and the reliability of physics-based, fine scale simulations, Kevrekidis et. al. proposed years ago an ``equation-free'' framework that effectively models coarse-scale behavior with appropriately initialized fine-scale simulations \cite{eqfree1, eqfree2, init_on_man_VANDEKERCKHOVE}.

In this paper, we propose a framework for the use of generative models as a tool that facilitates the initialization of simulations of deterministic dynamical systems (described by PDEs or ODEs) consistently on the low-dimensional manifold of the system -- and subsequently using simulations of the dynamical system initialized by the generative model to evolve along the manifold. Depending on the label used we can identify coarse-scale steady states, bifurcations, and/or the slow manifold in general.  The generative model of choice for the demonstrations in this work is a conditional SGM (cSGM), which relies on: (1) score matching, and (2) a training-free approximation of the score function \cite{liu2024diffusion}.


The paper is outlined as follows: Section \ref{sec:methods} provides an overview of the computational methods used throughout this work. Section \ref{sec:framework} lays out the proposed frameworks for using generative models to initialize on manifolds and identify steady states. Section \ref{sec:cusp} shows results pertaining to the use of the framework to identify bifurcations and steady states. Section \ref{sec:C-I} compares the proposed algorithms and demonstrates their uses applied to a reaction-diffusion PDE. Section \ref{sec:PFR} then demonstrates and discusses the application of the proposed framework to an example of a plug-flow tubular reactor. Section \ref{sec:conclusions} summarizes and discusses the results.

\section{Score-based Generative (Diffusion) Models (SGM) }
\label{sec:methods}
\noindent
SGMs are trained to generate new data points of a target density by utilizing a discretized stochastic differential equation (SDE)
that gradually transforms the data into normally distributed noise. This SDE takes the form of a reverse, or backward, SDE of a forward SDE of choice. The forward SDE is any SDE that gradually transforms data into normally distributed noise. In generic form, the forward SDE can be defined as

\begin{equation}
    \mbox{d} x = f(x,t) \mbox{d} t + g(t)\mbox{d} B_t,
    \label{eq:sde}
\end{equation}

where $f(x, t)$ and $g(x, t)$ are the drift and diffusion terms, respectively. The corresponding reverse-time SDE is:

\begin{equation}
    \mbox{d} x = [f(x,t) - g(t)^2\nabla_x \log p_t(x)]\mbox{d} t + g(t)\mbox{d} B_t.\\
    \label{eq:reversesde}
\end{equation}

\noindent
Because the end product of the forward SDE is normally distributed noise, we can initialize this reverse SDE with this distribution, and eventually obtain the original data distribution after integration. The initial and target distributions are known; however, the unknown of this problem is then the intermediate distributions and therefore the drift $\nabla \log p_t(x)$ (where $p_t(x)$ is the marginal distribution of $x$ at time $t$) of the reverse SDE, which we can then approach by e.g., training a neural network (whose inputs are the samples $x$ and respective times $t$, with learnable parameters denoted by $\theta$) to approximate the unknown $s_\theta(x,t) \approx \nabla \log p_t(x)$. The reverse SDE is implemented using the Euler-Maruyama method \cite{maruyama1955continuous}. To advance the SDE by $\delta t = t_{i+1} - t_i$, we use the following discretized equation:
\begin{equation}
    x(t_{i+1}) = x(t_i) + (\delta t) \left( f(x,t_i) - g(t_i)^2\nabla_x \log p_{t_i}(x)\right) + \sqrt{(t_{i+1} -  t_i)} g(t_i)B_{\delta t},
    \label{eq:euler_SGM_SDE}
\end{equation}
where $B_{\delta t}$ is a normal random variable with standard variance.
We have the discretized reverse SDE, so we then use a neural network to approximate $\nabla_x \log \hat{p}_t(x)$ (where $\hat{p}_t(x)$ refers to an approximation of the distribution ${p}_t(x)$). Next, we discuss different approaches for approximating the score function.

\subsection{Score-matching SGM}

\noindent
Score-Matching approaches \cite{songdiffmodels, songscore}, resulting in the following loss for the neural network approximation of the reverse SDE drift:
$${L}(\theta) = \mathbb{E}_{t \sim U(0,T)}\mathbb{E}_{x(0) \sim p(x(0))}\mathbb{E}_{x(t) \sim p(x(t) | x(0))}[\lambda (t) \| \nabla \log \hat{p}_t(x) - s_\theta(x(t), t)\|^2_2],$$
where $U(0,T)$ is a uniform distribution over the time interval of [0,T], $\lambda (t)$ is a positive weighting function, $x(0)$ denotes $x(t)$ at time $t=0$, and $\theta$ represents learned neural network weights. Note that this score-matching loss is comparable to a more computationally efficient \emph{denoising score matching objective} \cite{vincent2011connection}, which can be used as a substitute loss function. 

In this work, we are interested in generating conditional distributions, so we create a conditional diffusion model by having the neural network approximate $\nabla \log \hat{p}_t(x|y)$ where $y$ is some given label, using the following score-matching loss function, also called the conditional denoising estimator (CDE), discussed in other relevant work \cite{tashiro2021csdi, batzolis2021conditional}:
\begin{equation}
    {L}(\theta) = \mathbb{E}_{t \sim U(0,T)} \mathbb{E}_{x(0),y \sim p(x(0) | y)} \mathbb{E}_{x(t) \sim p(x(t) | x(0))}[\lambda (t)\| \nabla \log \hat{p}_t(x | y) - s_\theta(x(t), t, y)\|^2_2],
    \label{eq:cond_loss}
\end{equation}
\noindent meaning that the new reverse SDE that we will evaluate becomes:
\begin{equation}
    x({t_{i+1}}) = x({t_i}) + (t_{i+1} - t_i) \left( f(x,t_i) - g(t_i)^2\nabla_x \log p_t(x|y)\right) + \sqrt{t_{i+1} - t_i} g(t_i)B_{t_{i+1} - t_i}.
    \label{eq:cond_EM_SDE}
\end{equation}
This is essentially the same SDE as equation \ref{eq:euler_SGM_SDE} (the unconditional reverse SDE), but with a change to the Neural Net learned drift term. The final result is a neural-net based model that utilizes this discretized SDE to generate data consistent with target conditional distributions. All SDE parameters for our SGM models were chosen using guidance from Karras et. al. \cite{karrasdiffparams}, and further details regarding SGM training can be found in the appendix.

\subsection{Monte Carlo-based SGM}
\label{sec:MCS-SGM}

\noindent 
Several forms exist for the drift and diffusion of Eq.~(\ref{eq:sde}). Among these, specific parameterizations ensure that the terminal state is Gaussian \cite{song2020score, ho2020denoising, lu2022dpm}. Following the formulation proposed in \cite{liu2024diffusion}, we adopt the following definitions:
\begin{equation}
    f(x, t) = b(t) x_t = \frac{\text{d}\log \alpha_t}{\text{d}t}x_t
\end{equation}
and
\begin{equation}
    g(t) = \frac{\text{d}\beta_t^2}{\text{d}t} - 2\frac{\text{d}\log \alpha_t}{\text{d}t}\beta_t^2
\end{equation}
with $\alpha_t = 1-t$ and $\beta_t^2 = t$. For $t \in [0, 1]$ the SDE ensures a Gaussian terminal state \cite{song2020score}. 
In contrast to the previous discussed diffusion model where a trained neural network is utilized to estimate the score function $s(x, t) = \nabla_{x} \log p_t(x)$ of the reverse-time SDE in Eq.~(\ref{eq:reversesde}, the authors in \cite{liu2024diffusion}, proposed a mini-batch Monte Carlo Sampling (MCS) for estimating the score function. In this setting, the score function is approximated as:
\begin{equation}
s(x, t) \approx \sum_{n=1}^{N_m} \frac{x_t - \alpha_t x_j^n}{\beta_t^2} \frac{p(x_t|x_j^n)}{\sum_{m=1}^{N_m} p(x_t|x_j^m)} = \sum_{n=1}^{N_m} \frac{x_t - \alpha_t x_j^n}{\beta_t^2} \overline{w}_t(x_t, x_j^n)
\end{equation}
where
\begin{equation}
w_t(x_t, x_j^n) \approx \overline{w}(x_t, x_j^n) = \frac{p(x_t | x_j^n)}{\sum_{m=1}^N p(x_t | x_j^m)}
\end{equation}
are weights estimated by means of MCS; $\{x_{j}^{n}\}_{n=1}^{N_m}$ represents a mini-batch of the data, with $N_m \leq N$. It can be easily proven that  $p(x_t|x_0) \sim \mathcal{N}(\alpha_t x_0, \beta_t^2 \mathbf{I}_d)$ is the conditional Gaussian distribution. This approach utilizes a neural network to generate additional realizations: Starting with samples \( \{\textbf{y}_m\}_{m=1}^N \) drawn from \( \mathcal{N}(0, \textbf{I}_d) \), labeled data are generated which are utilized to train a neural network to learn the mapping $\mathcal{N}(0, \mathbf{I}_d) \to f_\textbf{x}(\cdot)$, by minimizing the mean squared error (MSE). Given that we are interested in generating  training input-output data for the neural network, we want the mapping \( \mathcal{N}(0, \textbf{I}_d) \rightarrow f_{\textbf{x}}(\cdot) \) to be ``smooth''. Therefore,  the method utilizes an ordinary differential equation (ODE) to approximate the reverse-time SDE:

\begin{equation}
\text{d}x_t = \left( b(t) x_t - \frac{1}{2} g(t) s(x, t) \right) \, \text{d}t
\end{equation}
After the neural network is trained, this method does not require solving the reverse-time ODE (or SDE) to generate samples of $f_{\textbf{x}}(\cdot)$, which makes the approach computationally very efficient.

Since in this work we are interested on sampling from conditional distributions, we can calculate the conditional score function as:

\begin{equation}
s(x,t, y)  = \sum_{n=1}^{N_m} \frac{x_t - \alpha_t x_j^n}{\beta_t^2} \cdot \frac{p(x_t | (x_j^n|y))}{\sum_{m=1}^N p(x_t | (x_j^m)|y)}
\end{equation}
where $x_j^n|y$ are mini-batch MCS samples conditioned to a label $y$.

\section{Manifold Learning}
\label{ssec:manifold}

\subsection{Diffusion Maps algorithm}
\label{ssec:dmaps}
\noindent
Diffusion Maps, introduced by Coifman and Lafon \cite{COIFMANdmaps} can reveal the \textit{intrinsic geometry} of a data set set \textbf{X} = $\{x_i\}^N_i$ sampled from a manifold $\mathcal{M}$ where each data point $\textbf{x}_i \in \mathbb{R}^m$.
This is accomplished by first constructing an affinity matrix $\textbf{A} \in \mathbb{R}^{N \times N}$. The entries of $\textbf{A}$ are computed with a kernel function. In our case, we use the Gaussian kernel,
\begin{equation*}
    A_{ij} = \exp \bigg(-\frac{\| x_i - x_j \|^2}{2\epsilon}\bigg),
\end{equation*}
where $||\cdot||$ denotes a norm (in this work, we use the Euclidean norm) and $\epsilon$ is the kernels' parameter that determines the rate of decay of the kernel. A normalization is applied to $\textbf{A}$ that allows us to compute the intrinsic geometry of $\textbf{X}$ independent of its sampling density 

\begin{equation*}
    \Bar{\textbf{A}} = \textbf{P}^{-1}\textbf{A}\textbf{P}^{-1}
\end{equation*}
where the entries of matrix $\textbf{P}$ are computed as $
    P_{ii} = \sum_{j=1}^{N} A_{ij}$.
A second normalization is applied to matrix $\Bar{\textbf{A}}$ 
\begin{equation*}
    W_{ij} = \frac{\Bar{A}_{ij}}{\sum_{j=1}^{N} \Bar{A}_{ij}}.
\end{equation*}
to construct the Markovian matrix
 $\textbf{W}$.
 The eigendecomposition of $\textbf{W}$ is then computed to get a set of eigenvectors $\phi$
 A proper selection of the eigenvectors that parameterize independent directions, \textit{non-harmonic} eigenvectors is then applied. This can be achieved by visual inspection for dimensions up to three (see the Appendix in \cite{evangelou2023double}). For dimensions larger than three, a local linear regression algorithm can be used for an automatic selection of the non-harmonic eigenvectors \cite{dsilva2018parsimonious}. In our case the local linear regression algorithm was used through the Python implementation in the Python library \textit{datafold} \cite{Lehmberg2020}. If the number of the obtained non-harmonic eigenvectors is smaller than the original dimensions of the data $m$, dimensionality reduction has been achieved. In our work, the Diffusion Maps algorithm was used to discover a set of reduced variables, which we use to condition the cSGM in order to generate points on the manifold (see Section \ref{sec:C-I}).

\subsection{Lifting Operations using Geometric Harmonics}
\label{ssec:GH}

\noindent
 Geometric harmonics (GH) introduced also by Coifman and Lafon \cite{coifman2006geometric} is a scheme  based on the Nystr\"om Extension formula \cite{Nystrom_Extension} that extends a function $f$ defined on a data set $\textbf{X}$ sampled from a manifold $\mathcal{M}$ for out-of-sample data points $\vect{x}_{new}\notin\textbf{X}$. 
The first step for the GH algorithm, as in Diffusion Maps, is to  construct an affinity matrix $\textbf{A}^{\star}$

\begin{equation*}
    A^{\star}_{ij} = \exp \bigg(-\frac{\| x_i - x_j \|^2}{2\epsilon^{\star}}\bigg).
\end{equation*}

\noindent
Note that the hyperparameter $\epsilon^{\star}$ in practice might be different in this case because we wish to construct a regression scheme rather than discover the intrinsic geometry of the data.

We then compute the eigendecomposition of $\textbf{A}^{\star}$. The
matrix $\textbf{A}^{\star}$ has a set of orthonormal vectors $\psi_0,\psi_1,\ldots,\psi_{N-1}$ and non-negative eigenvalues ($\sigma_0\geq\sigma_1\geq\cdots\geq\sigma_{N-1}\geq0$).  Compute a subset of these eigenvectors, for $\delta>0$ $S_{\delta}=\{\alpha\,:\,\sigma_{\alpha}>\delta\sigma_{0}\}$. 
This (truncated) set of eigenvectors is used as the basis in which we project the sampled points of $f$.

\begin{equation*}
    f \mapsto P_{\delta}f =  \sum_{\alpha \in S_{\delta}} \langle f,\psi_{\alpha} \rangle \psi_{\alpha}, 
\end{equation*}

\noindent
where $\langle \cdot, \cdot \rangle$ denotes the inner product. This project is performed only once, it can be viewed as the training step of the GH algorithm.

 The extension of $f$ for a new point $\vect{x}_{new}\notin\textbf{X}$ is defined as
\begin{equation}
\label{eqn:Geometric_Harmonics_Extension}
    (Ef)(\vect{x}_{new}) = \sum_{{\alpha\in{S_{\delta}}}} \langle f,\vect{\psi}_{\alpha}\rangle\vect{\Psi}_{\alpha}(\vect{x}_{new})\,.
\end{equation}  
\noindent
where
\begin{equation}
\Psi_{\alpha}(\vect{x}_{new}) = \sigma^{-1}_{\alpha} \sum_{j=1}^N \exp \bigg(-\frac{\| x_{new} - x_j \|^2}{2\epsilon^{\star}}\bigg)\vect{\psi}_{\alpha}(\vect{x}_j).
\label{eq:GH_extension}
\end{equation}

\noindent
In Equation \eqref{eq:GH_extension} $\psi_{\alpha}(\vect{x}_j)$ is the $j^{\text{th}}$ components of the eigenvector $\psi_{\alpha}$. In this work, Geometric Harmonics are used to construct mappings from a dataset with reduced dimensions to the original dataset in ambient space. Utilizing this mapping can reduce the impact of noise on the data sampled by the SGM (see Algorithm \ref{alg:Training_SGM_alg} and Section \ref{sec:C-I}).


\section{Proposed Data-driven Framework}
\label{sec:framework}

\noindent
The workflow in Algorithm \ref{alg:Training_SGM_alg} below can be used to perform both tasks outlined in Section \ref{sec:intro}: (a) the initialization on slow manifolds consistent with previous observations as well as (b) initialization on a bifurcation surface consistent with steady states observed at previously sampled parameter values. Interestingly, if the steady states in question are steady states of deterministic \textit{partial differential equations}, the same approach can be used to reconstruct {\em full spatial profiles} of the steady states on the bifurcation surface from partial previous local spatial data observations: one can ``lift'' from partial previous observations to the full ambient space. The output of the trained cSGM and these reconstructed profiles can then be used to reconstruct steady state profiles at previously unobserved parameter values for a system of PDEs.

\begin{algorithm}[H]
\caption{Training a conditional SGM to initialize on a slow manifold}
{\textbf{Input:} Samples from a dynamical system, $x(t)$, after discarding transient data, assumed to lie near a reduced-dimension slow or bifurcation manifold; corresponding distribution $P(x)$. 
and labels $y$ as selected Quantities of Interest (QoI) that parametrize the manifold.}
\label{alg:Training_SGM_alg}
\begin{algorithmic}[0]
\State Select QoI(s), $y$, either through \emph{a priori} knowledge or using nonlinear dimensionality reduction techniques
\If{Dimensionality reduction is possible}
    \State Truncate $x(t)$ to a lower-dimensional dataset $x_r(t)$ that includes the QoI(s)
    \State Sample from $x_r(t)$
    \State Train a cSGM on $(x_r, y)$ pairs to model the conditional distribution over the manifold
    \State Use the trained cSGM to sample from $\hat{P}(x_r \mid y)$ — approximated conditional measures in reduced space
    \State  Use \textit{Geometric Harmonics} to learn a mapping from $x_r(t)$ to $x(t)$
    \State Apply \textit{Geometric Harmonics} to lift generated samples $\hat{P}(x_r \mid y)$ to an estimate of $\hat{P}(x \mid y)$ \\
\Else
    \State Sample from $x(t)$
    \State Train a cSGM on $(x, y)$ pairs to model the conditional distribution over the manifold
    \State Use the trained cSGM to sample from $\hat{P}(x \mid y)$ — approximated conditional measures in ambient space \\
\EndIf \\
\Return $\hat{P}(x \mid y)$ evaluated at the given $y$ values
\end{algorithmic}
\end{algorithm}

As expected, sampling data on the manifold with a SGM introduces noise into the samples due to the stochastic nature of the generative model. Thus, if one wishes to minimize the impact of the introduced noise on the generated data, it is important to consider the dimensionality of the data provided to the generative model. Thus, Algorithm \ref{alg:Training_SGM_alg} is presented as a method to minimize the noise introduced to the generated samples by using Geometric Harmonics in tandem with SGMs.

\section{Numerical Examples}
\noindent
In this section we demonstrate the efficiency of the proposed methods to reconstruct steady-state bifurcation diagrams/surfaces for various systems.

\subsection{Example 1: Using cSGMs to reconstruct steady state bifurcation diagrams/surfaces}
\label{sec:cusp}
\noindent
The first example demonstrates the use of SGMs for sampling bifurcation surfaces and steady states on them consistent with labels (parameters, or system steady state values). Consider a simple cusp bifurcation surface in $x, \lambda, \mu$ space given by the equation
\begin{equation}
    \dot{x} =-x^3 + \lambda x + \mu
\end{equation}
where the manifold is parameterized by the variable $x$ and the two control parameters $\lambda$ and $\mu$. The training dataset provided to the cSGM consists of $20,000$ samples, where $x$ is uniformly sampled in the range $[-2, 2]$, and $\lambda$ in the range $[-2.5, 2.5]$. The corresponding values of $\mu$ are determined by solving the relation $\mu = x^3 - \lambda x$. This dataset, which lies on the bifurcation surface, is used to train the cSGM. Once trained, the cSGM enables the generation of steady-state and parameter values on the manifold, conditioned on \emph{new} out-of-sample state and parameter inputs. In this section, we explore two distinct cases for training the generative model:

\begin{enumerate}
    \item Case 1: training a cSGM using the value of one of the control parameters of the cusp as a label, $\mu$; this produces distributions of data on the bifurcation surface that are consistent with the one-parameter bifurcation diagram along the second control parameter, $\lambda$.
    \item Case 2: training a cSGM with a label consisting of {\em  both} control parameter values, $\mu$ and $\lambda$, producing distributions of data in the form of clusters at the steady states of the system at the prescribed parameter values.
\end{enumerate}

 \begin{figure}[!htb]
     \centering
     \includegraphics[width=0.4\linewidth]{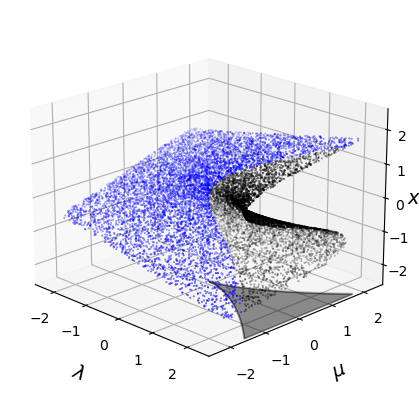}
     \caption{Data produced by the cusp bifurcation equations shown as points with its 2D projection to the control parameter space below the samples. The black points (and shaded region of the projection) correspond to the values of $\mu$ and $\lambda$ for which there exist three steady states, and the blue points (and unshaded region of the projection) correspond to the parameter values at which there is a single steady state.}
     \label{fig:cusp_training_data}
 \end{figure}

\begin{figure}[H]
     \centering
     \includegraphics[width=1.0\linewidth]{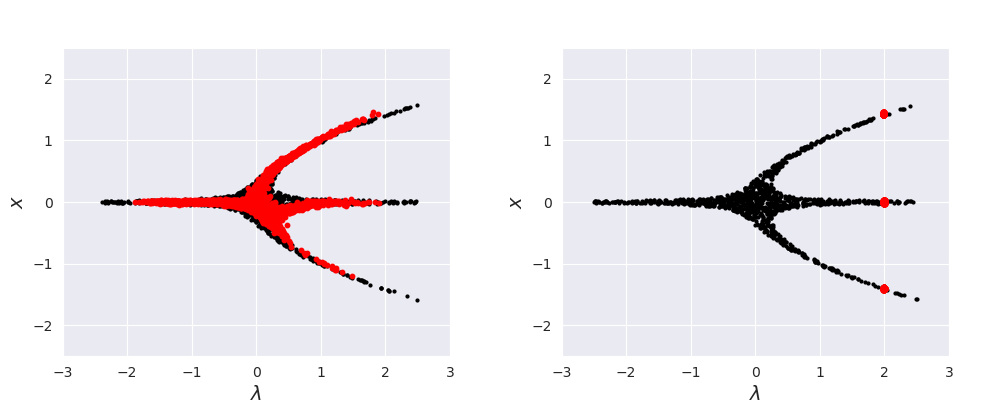}
     \caption{cSGM: A slice of training data where $\mu \approx 0$ (black points) plotted with three-dimensional outputs from a cSGM (red points) where the conditioned values are $\mu = 0$ (left) and $\mu = 0, \lambda = 2$ (right).}
     \label{fig:cusp_slices}
 \end{figure}

Figure \ref{fig:cusp_training_data} shows the training data sampled from the manifold of the cusp, along with a shaded region of steady state multiplicity in control parameter space.  Figure \ref{fig:cusp_slices} shows a slice of the training data where $\mu \approx 0$, compared with the output of a cSGM conditioned on a single parameter value ($\mu = 0$ in the left plot) and conditioned on vlaues of both parameters ($\mu = 0$ and $\lambda = 2$ in the right plot). One can clearly see that there is a single steady state is generated when $\lambda < 0$ and three steady states when $\lambda > 0$. 
Figure \ref{fig:cusp_slices} demonstrates that the cSGM can produce reasonable initializations on the manifold of the cusp. This manifold is three-dimensional, and when given a one-dimensional label, the cSGM generates a family of values in the space of the remaining parameter and the state variable close to the two-dimensional bifurcation surface in the  three-dimensional embedding space, effectively sampling/reconstructing the bifurcation diagram for the fixed label value. Notice however the ``fatness" of the reconstructed curves/points - a subject to which we will return later in the manuscript.
When given a two-dimensional label, the cSGM produces a family of scalar values  of the remaining manifold coordinate; this effectively takes the form of ``bloblike" clusters of data containing the steady states of the system for the prescribed parameters. In the case shown in Figure \ref{fig:cusp_slices}, we can see three distinct clusters of generated points around the relevant values of $x$ (since our conditioning values of $\mu$ and $\lambda$ lie in the multiplicity region).
%

\begin{figure}[H]
     \centering
     \includegraphics[width=1.0\linewidth]{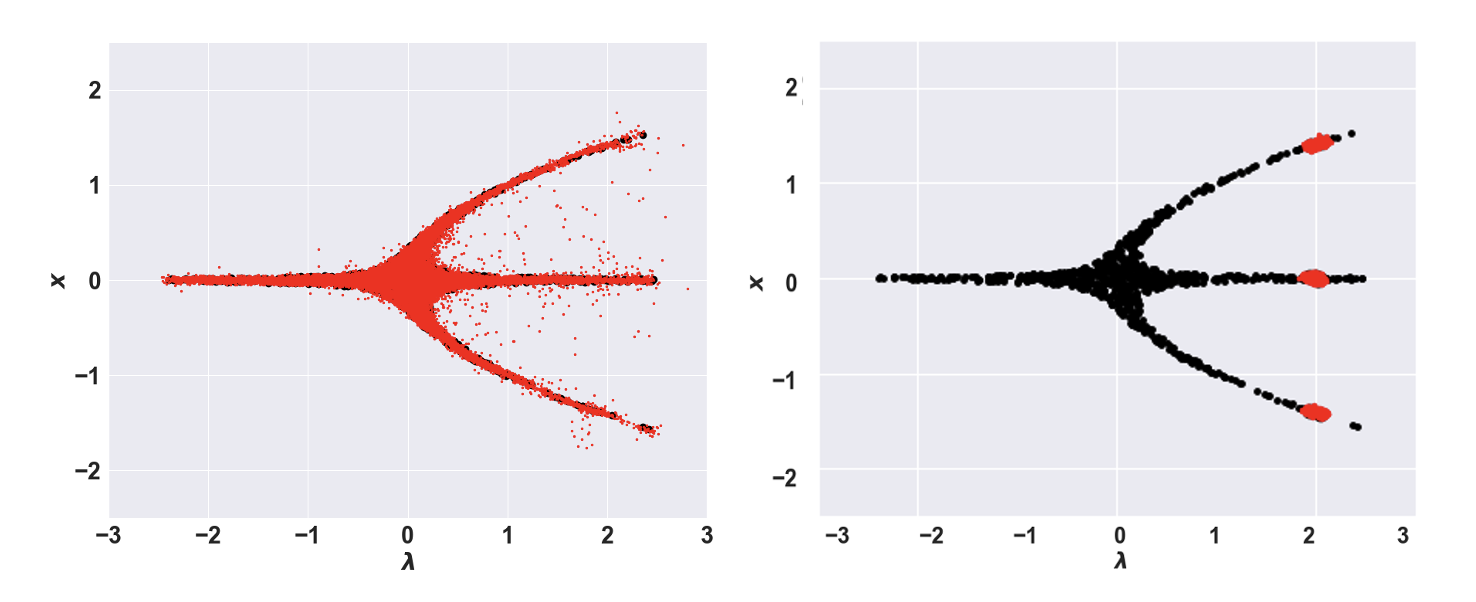}
     \caption{MCS-based cSGM: A slice of training data where $\mu \approx 0$ (black points) plotted with three-dimensional outputs from the MCS-based cSGM (red points) where the conditioned values are $\mu = 0$ (left) and $\mu = 0, \lambda = 2$ (right).}
     \label{fig:cusp_slices_MCS}
 \end{figure}

Figure \ref{fig:cusp_slices_MCS}  shows that the MCS-based cSGM can generate meaningful samples on the cusp manifold. With a one-dimensional label, it reconstructs a slice of the bifurcation surface in the 3D embedding space, approximating the underlying structure. When conditioned on two parameters, the MCS-based cSGM produces clusters of scalar outputs for the remaining coordinate, capturing the distribution of steady states. Again, three distinct clusters align with the multiplicity region defined by the specified $\mu$ and $\lambda$ values. The spread of the generated samples reflects the fact that we are drawing from a learned probability density using Monte Carlo simulation, a fundamentally stochastic process. While the true spread of the underlying distribution remains fixed, the observed spread in the samples can vary depending on the number of samples used. With fewer samples, the estimated spread may appear noisy or imprecise due to sampling variability. As the number of Monte Carlo samples increases, the empirical distribution more accurately captures the true variability of the learned density, leading to a clearer and more stable representation of its spread. To overcome this challenge, in a recent work, the authors proposed a method for generative learning of densities on manifolds \cite{giovanis2025generative}.


\subsection{Example 2: cSGM initialization on the inertial manifold of the Chafee-Infante PDE}
\label{sec:C-I}
\noindent
In this example we generate data using a trained cSGM to initialize on the inertial manifold of the Chafee-Infante PDE (an inertial manifold is a "slow" invariant manifold that is exponentially attracting). As stated in Section \ref{sec:framework}, we can use an algorithm (Algorithm \ref{alg:Training_SGM_alg})
to reduce the impact of the noise introduced to the system by the SGM. To compare the proposed method within algorithm \ref{alg:Training_SGM_alg} with using the SGM without dimensionality reduction, consider the Chafee-Infante reaction diffusion equation


\begin{equation}
    u_t = u - u^3 + \nu u_{xx}
\end{equation}
with $\nu = 0.16$ and boundary conditions $u(0,t) = u(\pi,t) = 0$. 
For this PDE it has been shown that at this value of $\nu$ the long-term dynamics live in a two-dimensional manifold \cite{CI_slow_man_int, CI_jolly}. To reveal this 2-D manifold (that is embedded in a higher dimensional space) one can use nonlinear dimensionality reduction techniques. First to sample the data, we approximate $u$ with a discrete sine transformation

\begin{equation}
    u(x,t) \approx \sum_{k=1}^{10} \alpha_k(t) \sin{(kx)}. 
\end{equation}

The Galerkin projection onto these leading ten Fourier modes results in a system of (spectral) differential equations that take the form of $\frac{d \alpha}{dt} = f(\alpha)$,
where $\alpha \in \mathbb{R}^{10}$. It has been established that these ten modes provide a (for all practical purposes) converged discretization of the problem. Following the procedure proposed in Section \ref{sec:framework}, we discard the transient data and keep only ``snapshots" of the long term dynamics. Clearly, in this case, the embedding dimension is 10 (the ten axes are the coefficients of the snapshot data on each of the ten Fourier modes), and the (approximate, slow) inertial manifold is a two-dimensional manifold embedded in this 10-D space. Integrating this system of ODEs at various initial conditions ($\alpha_{t_0}$) and discarding the transients resulted in a training data set of ~3600 data points.
To establish the two-dimensionality of the slow manifold in a data-driven way we use the  nonlinear dimensionality reduction technique Diffusion Maps \cite{COIFMANdmaps}. We then observe that the two leading diffusion map coordinates $\phi_1, \phi_2$ that parameterize the 2-D manifold are one-to-one with the first two Fourier modes, $\alpha_1$ and $\alpha_2$ (there exists a smooth invertible transformation between the two pairs, whose Jacobian determinant is bounded away from zero on the data, see Section \ref{ssec:dmaps} on how these calculations are performed and see \cite{evangelou2023double} for a demonstration of this with the Chafee-Infante). Either pair parametrizes the slow manifold, and we can take measurements from either pair (the {\em a priori} theoretical known $a_i$ or the data-driven $\phi_i$) as our labels - on which to condition the cSGM so that it can generate consistent full solution profiles on the approximate inertial manifold. 


\subsubsection{Case I: A Comparison of Algorithms using \emph{a priori} Known Labels}

\noindent
Selecting one of the known parameters of the manifold as a label (here $\alpha_1$), we then train the cSGM on the data consisting of the the 10
Fourier modes or train a separate cSGM on the first two Fourier modes (Algorithm \ref{alg:Training_SGM_alg}). Using the notation of the algorithm, $y=\alpha_1$ for both cases. $P(x) = \alpha_1 - \alpha_{10}$ for the cSGM, and $P(x) = \alpha_1,\alpha_2$ for Algorithm \ref{alg:Training_SGM_alg}.

\begin{figure}[!htb]
    \centering
    \includegraphics[width=0.8\linewidth]{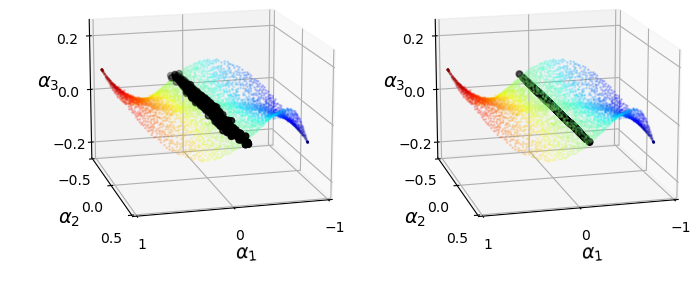}
    \caption{In both plots, the first three Fourier modes, $\alpha_1$, $\alpha_2$, and $\alpha_3$ are plotted on the x, y, and z axes respectively and are colored by $\alpha_1$. On the left, a 10-D (all ten Fourier modes) cSGM output of 1000 samples conditioned at $\alpha_1 = 0$ is plotted over the manifold as black points. On the right, a 2-D (only the first two Fourier modes) cSGM output of 1000 samples conditioned at $\alpha_1 = 0$ is plotted over the manifold as black points.}
    \label{fig:CI_3d_comparison}
\end{figure}
\begin{figure}[!htb]
    \centering
    \includegraphics[width=0.8\linewidth]{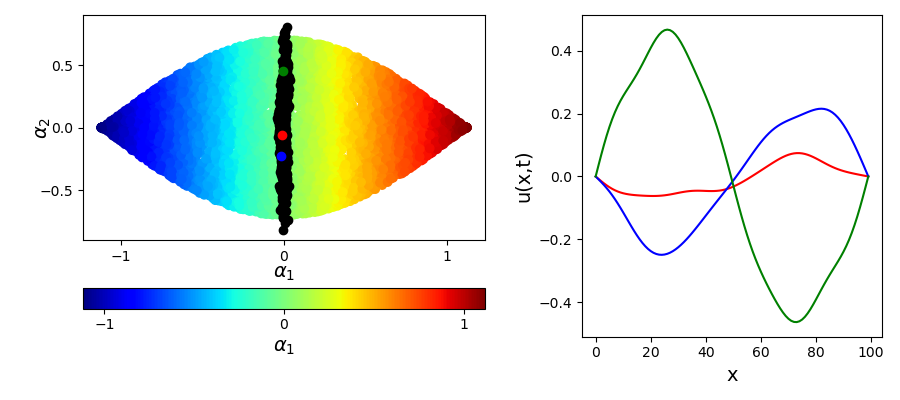}
    \caption{On the left, the first two Fourier modes $\alpha_1$ and $\alpha_2$ are plotted on the x and y axes respectively and are colored by $\alpha_1$. A cSGM output of 1000 samples of all ten Fourier modes conditioned at $\alpha_1 = 0$ is plotted over the manifold as black points. Three points from this generated set are colored red, blue, and green. On the right, three reconstructed $u(x,t)$ profiles corresponding to the three colored dots in the left plot are shown.}
    \label{fig:CI_with_modes_10D}
\end{figure}

\begin{figure}[!htb]
    \centering
    \includegraphics[width=0.8\linewidth]{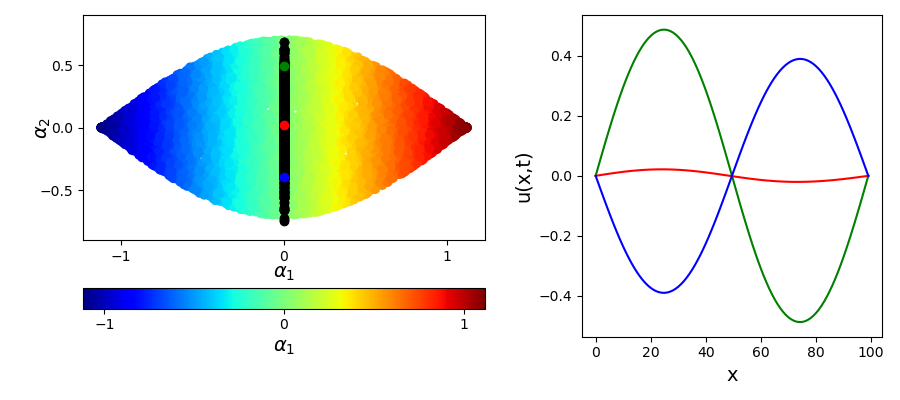}
    \caption{On the left, the first two Fourier modes $\alpha_1$ and $\alpha_2$ are plotted on the x and y axes respectively and are colored by $\alpha_1$. A cSGM output of 1000 samples of the first two Fourier modes conditioned at $\alpha_1 = 0$ is plotted over the manifold as black points. Three points from this generated set are colored red, blue, and green. On the right, three reconstructed $u(x,t)$ profiles corresponding to the three colored dots in the left plot are shown.}
    \label{fig:CI_with_modes_2D}
\end{figure}

\begin{figure}[!htb]
    \centering
    \begin{subfigure}[b]{0.33\textwidth}
        \centering
        \includegraphics[width=\textwidth]{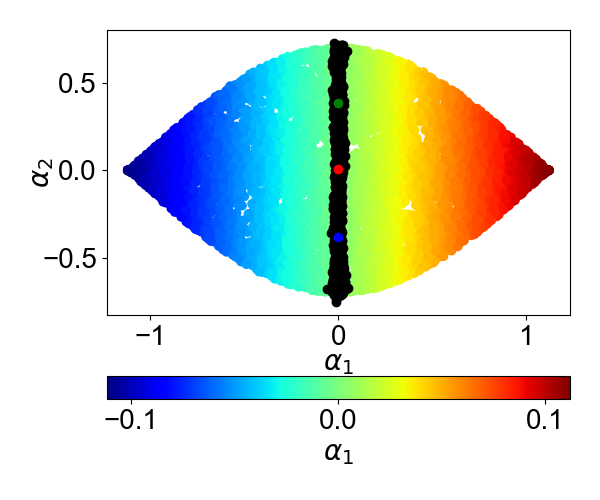}
        \caption{}
        \label{fig:sub3}
    \end{subfigure}
    \hfill
        \begin{subfigure}[b]{0.33\textwidth}
        \centering
        \includegraphics[width=\textwidth]{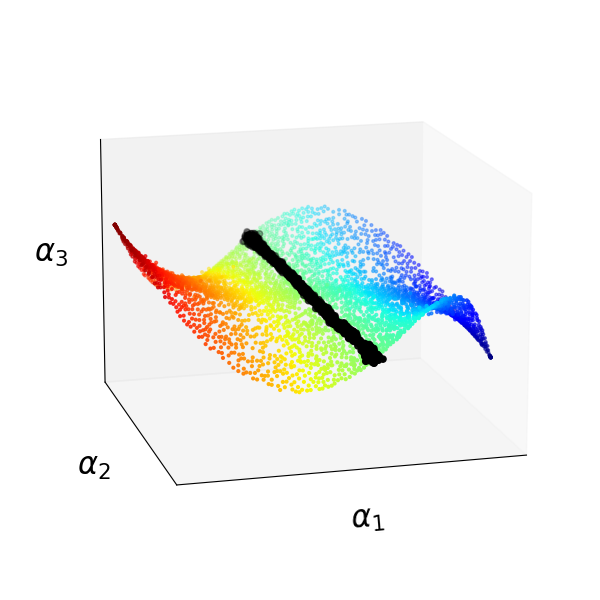}
        \caption{}
        \label{fig:sub2}
    \end{subfigure}
    \hfill
    \begin{subfigure}[b]{0.33\textwidth}
        \centering
        \includegraphics[width=\textwidth]{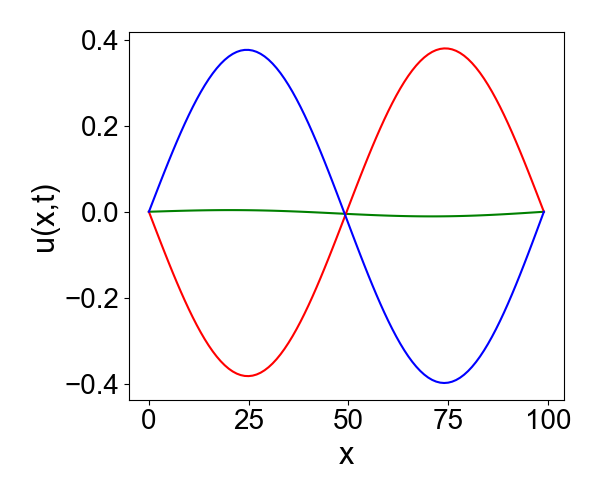}
        \caption{}
        \label{fig:sub4}
    \end{subfigure}
    \caption{Conditional MCS-based SGM. Case: 10 Fourier modes conditioned at $\alpha_1=0$. Three points from this generated set are colored red, blue, and green. On the right, three reconstructed $u(x,t)$ profiles corresponding to the three colored dots in the left plot are shown.}
    \label{fig:MCS-CI_with_modes_2D}
\end{figure}

Figure \ref{fig:CI_3d_comparison} shows that a generative model can produce values that are consistent with a particular conditional distribution on the manifold when generating either ten or two Fourier modes (with noise visually present in the ten mode generation). However, the differences in the output of the two methods becomes apparent when using the Fourier modes to produce the original function profiles of the PDE. Figure \ref{fig:CI_with_modes_10D} shows that the noise present in the samples produced by the SGM causes the reconstructed profiles to be less smooth than the functions produced by Algorithm \ref{alg:Training_SGM_alg} in Figure \ref{fig:CI_with_modes_2D}. The reconstruction of the remaining eight modes with Geometric Harmonics results in a minimization of noise added to the sampling of the system, and produces smooth function profiles. Figures \ref{fig:MCS-CI_with_modes_2D} shows the outputs of the MCS-based cSGM. Compared to standard cSGM samples, the generated points exhibit a wider spread, reflecting the inherent variability introduced by the noise of the MCS sampling process and the in the MCS prediction NNs.


\subsubsection{Case II: Initialization in the manifold, with labels identified through nonlinear dimensionality reduction}

\begin{figure}[!htb]
    \centering
    \begin{subfigure}[b]{0.49\textwidth}
        \centering
        \includegraphics[width=\textwidth]{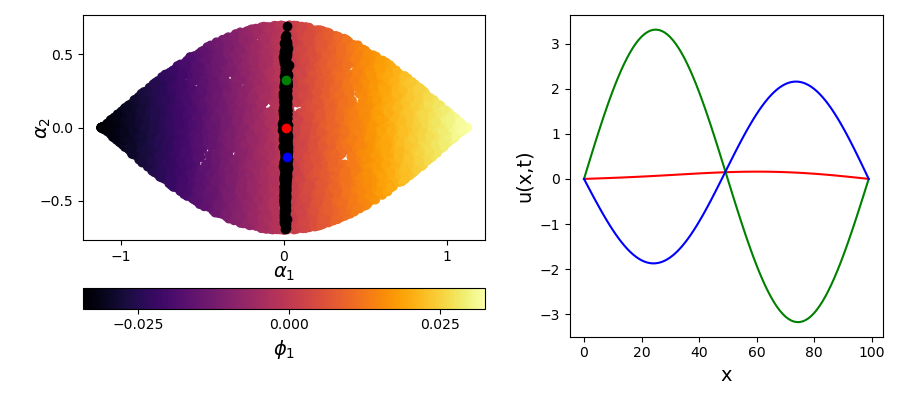}
        \caption{}
        \label{fig:sub5}
    \end{subfigure}
    \hfill
        \begin{subfigure}[b]{0.49\textwidth}
        \centering
        \includegraphics[width=\textwidth]{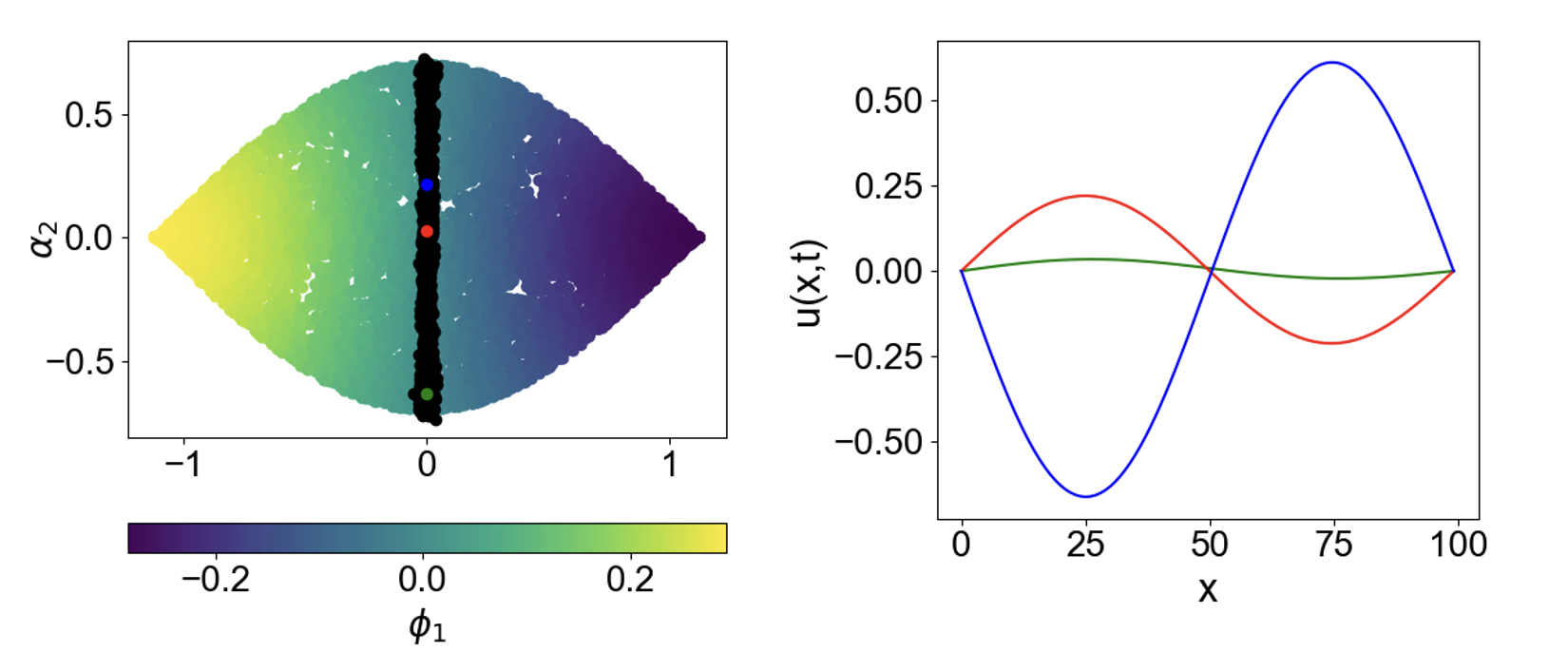}
        \caption{}
        \label{fig:sub6}
    \end{subfigure}
    \caption{(a) On the left, the first two Fourier modes $\alpha_1$ and $\alpha_2$ are plotted on the x and y axes respectively and are colored by the diffusion map coordinate, $\phi_1$, that is one-to-one with $\alpha_1$. A cSGM output of 1000 samples conditioned at $\phi_1 = 0$ is plotted over the manifold as black points with three sample points taken from the set and colored red, blue, and green. On the right, the three reconstructed $u(x,t)$ profiles corresponding to the three colored samples in the left plot are shown. (b) The corresponding MCS-based cSGM samples (black points) conditioned on $\phi_1 = 0$, together with the three reconstructed $u(x,t)$ profiles.}
    \label{fig:cSGM_output_on_CI_dmaps}
\end{figure}

\noindent
As detailed in Section \ref{ssec:dmaps} and \ref{sec:C-I}, we can use diffusion maps to identify coordinates that are one-to-one with the Fourier modes that parametrize the manifold. Using one of these coordinates as a label for the cSGM (here, the coordinate $\phi_1$ is one-to-one with $\alpha_1$), we train the cSGM so that it can generate consistent full solution profiles on the approximate inertial manifold. Figure \ref{fig:cSGM_output_on_CI_dmaps} demonstrates the use of Algorithm \ref{alg:Training_SGM_alg} to generate samples of the first two Fourier modes consistent with the prescribed value of the label $\phi_1$ then reconstruct the remaining eight modes with Geometric Harmonics. In \cite{giovanis2025generative}, the authors proposed a slight tweak to this framework: the double diffusion map method proposed by Evangelou et. al. \cite{evangelou2023double} was used to reconstruct samples in the original ambient space from the diffusion map coordinate space. This allows the cSGM to be trained to generate samples in diffusion map coordinate space rather than Fourier mode space.

\subsection{Example: cSGM initialization of a plug flow reactor PDE system}
\label{sec:PFR}
\begin{figure}[!htb]
    \centering
    \includegraphics[width=0.5\linewidth]{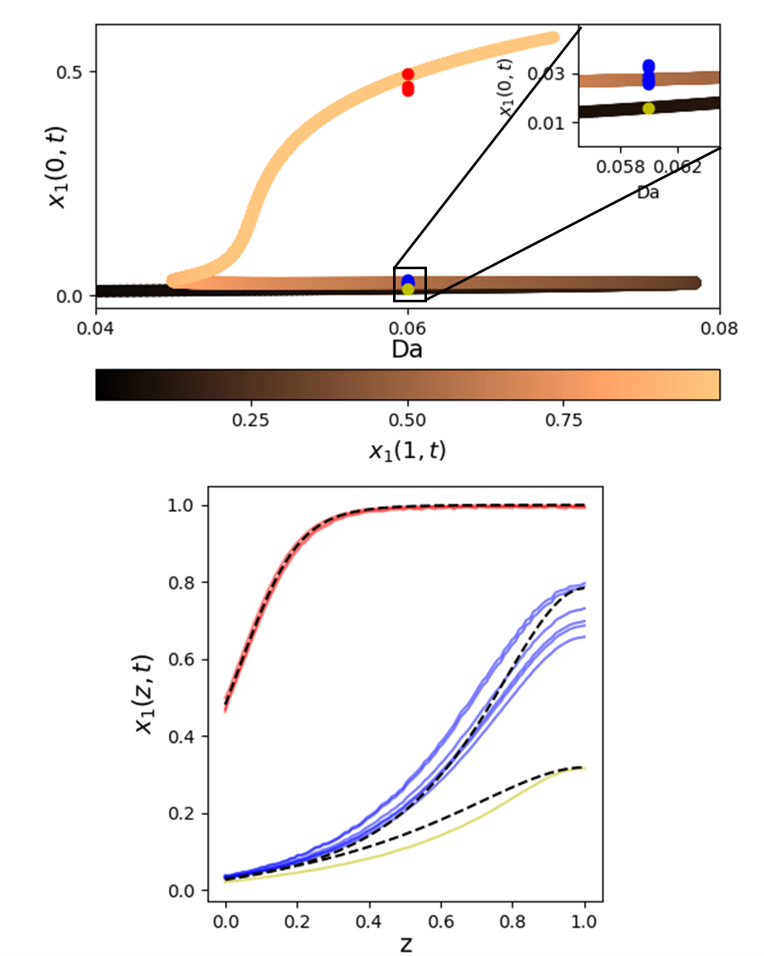}
    \caption{Above: training data of initial conversion values of the PFR ($x_1$ at $z = 0$) are plotted for varying Da values shaded in grayscale by the conversion value at the end of the respective profile ($x_1$ at $z = 1$). Four samples from a SGM output conditioned at $\mathrm{Da} = 0.06$ are plotted over the training data as individual points of varying colors with an additional closer view of the lower branches and the generated points on the upper branch. Below: profiles of conversion values across the length of the reactor that have been reconstructed from the generated initial conversions are plotted in the same respective color as the initial condition. Three profiles from the training data (one for each general cluster of points) are plotted as black dashed lines for reference.}
    \label{fig:cSGM_PFR_conversions_Da06_with_profiles}
\end{figure}

\begin{figure}[!htb]
    \centering
    \includegraphics[width=0.5\linewidth]{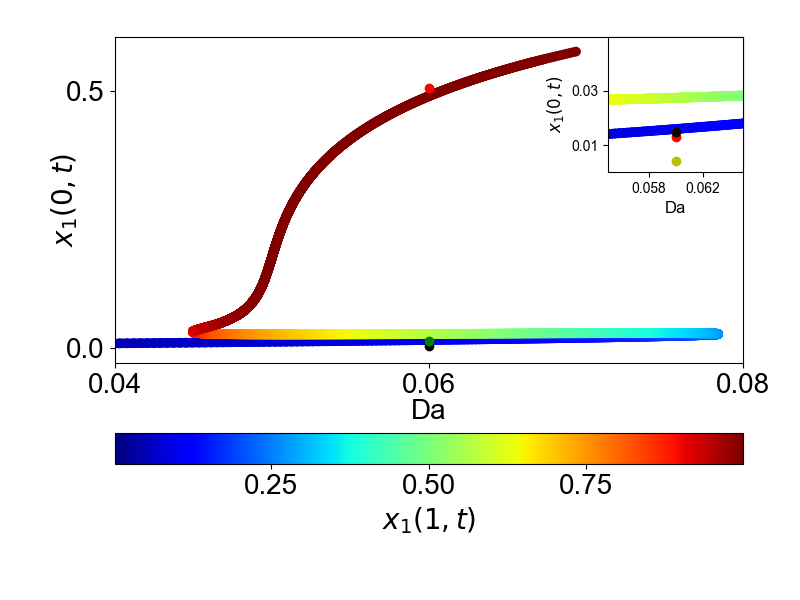}
    \caption{Four samples from the conditional MCS-based SGM output conditioned at $\mathrm{Da} = 0.06$ are plotted over the training data as individual points of varying colors with an additional closer view of the lower branches and the generated points on the upper branch.}
    \label{fig:MCS-cSGM_PFR_conversions_Da06_with_profiles}
\end{figure}

\noindent
As a more complex example, consider a first order irreversible reaction in a plug flow (turbulent) tubular reactor where the pseudohomogenous axial dispersion model is described by two coupled nonlinear dimensionless parabolic PDEs \cite{JENSEN1982199}:
\begin{equation}
    \begin{split}
        \frac{\partial x_1}{\partial t} &= \frac{1}{\mathrm{Pe}} \frac{\partial^2 x_1}{\partial z^2} - \frac{\partial x_1}{\partial z} + \mathrm{Da}(1 - x_1)\exp \left( \frac{x_2}{1+\frac{x_2}{\gamma}} \right) \\
        \frac{\partial x_2}{\partial t} &= \frac{1}{\mathrm{Pe}} \frac{\partial^2 x_2}{\partial z^2} - \frac{\partial x_2}{\partial z} - \beta x_2 + \mathrm{B Da}(1 - x_1)\exp \left( \frac{x_2}{1+\frac{x_2}{\gamma}} \right)
    \end{split}
\end{equation}
where $z=x/L$ with $0 < z < 1$. $x_1$ being the conversion and $x_2$ is the dimensionless temperature. B is the dimensionless adiabatic temperature rise, Da is the Damk\"ohler number, $\beta$ is a dimensionless heat transfer coefficent, Pe is the P\'eclet number, $\gamma$ is the ratio of the activation energy over the ideal gas constant $R$ and a reference temperature (i.e. $\gamma = E_a/RT_0$). The boundary conditions are defined as: 
\begin{equation*}
    \genfrac[]{0pt}{}{\frac{\partial x_1}{\partial z} =  \mathrm{Pe}x_1}{\frac{\partial x_2}{\partial z} =  \mathrm{Pe}x_2} z = 0,
    \genfrac[]{0pt}{}{\frac{\partial x_1}{\partial z} =  0}{\frac{\partial x_2}{\partial z} =  0} z = 1.
\end{equation*}
The training data for this example was generated using integration schemes from Koronaki et. al. \cite{KORONAKI2003951}. The fixed parameters for the experiment shown are $\mathrm{B} = 12$, $\beta = 0.5$, and $\gamma = 20$. Pe was also fixed at Pe = 5, while Da was used as the label given to the cSGM. For these PDEs, we can train a cSGM to produce initial conversion $x_1$ and dimensionless temperature $x_2$ values for given values of the parameters Da and Pe. Figure \ref{fig:cSGM_PFR_conversions_Da06_with_profiles} shows initial conversion values conditioned on Da values for a fixed Pe value, where a bifurcation occurs approximately where $\mathrm{Da} \approx 0.045$. With the produced initial conversion values, we can then reconstruct full conversion profiles across the length of the reactor, $z$, with Geometric Harmonics. It follows that it is possible to also generate dimensionless temperature values and subsequently temperature profiles along the length of the reactor also for unsampled spatial locations. Furthermore, if data is produced for an unfixed Da and Pe instead of fixing the value of one parameter, one could condition on Pe values as well as both Pe and Da values. This does require significantly more computational time to produce the training data required when both parameters are free; so in this work we focus on studies where the Peclet number remains fixed; the two-free parameter case is, as we noted, conceptually straightforward, but computationally costly. Figure \ref{fig:MCS-cSGM_PFR_conversions_Da06_with_profiles} depicts four samples from the conditional MCS-based SGM output conditioned at $\mathrm{Da} = 0.06$.

\subsection{Sampling on Non-uniformly Sampled Manifolds}

\begin{figure}[!htb]
    \centering
    \includegraphics[width=0.5\linewidth]{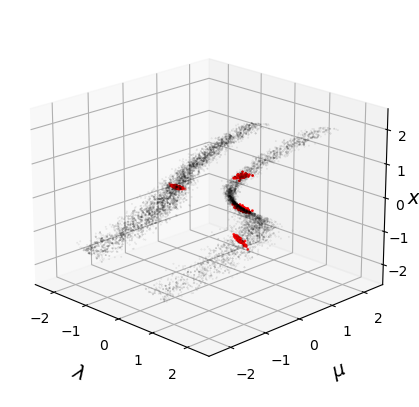}
    \caption{Training data produced by the cusp bifurcation equation shown as gray points with a bimodal distribution now present in the variable $\lambda$ compared with red points generated by the cSGM conditioned at $\mu = 0$.}
    \label{fig:nonuni_training_data}
\end{figure}

\noindent
During the collection of data/observations of dynamical systems and their underlying manifolds, uniformly sampling the system is neither a requirement nor a guarantee. 
As such, the distribution of the samples on the manifold may be an additional feature of interest. Generative models (such as the cSGMs discussed in this work) learn or approximate the distributions of their training data, so that, if the distributions of the training data are also of interest, the conditional distributions of a cSGM trained on the data can be observed. Consider the bifurcation example given in section \ref{sec:cusp}:
\begin{equation}
    \dot{x} =-x^3 + \lambda x + \mu
\end{equation}
where again the manifold is parameterized by the variable $x$ and the two control parameters $\lambda$ and $\mu$. Similarly to the example given in Section \ref{sec:cusp}, the training data given to the cSGM consists of 20,000 samples where $x$ ranges from -2 to 2, $\lambda$ ranges from -2.5 to 2.5, and the values of $\mu$ that correspond to these values of $x$ and $\lambda$ are found by solving $\mu = x^3 - \lambda x$. However, we now add an additional constraint: the data in $\lambda$ are sampled according to a bimodal distribution that is a mixture of two normal distributions, with the two modes being $\lambda = -1, 1$ and the variance of the normal distributions centered at these two modes being $\sigma = 0.5$. Figure \ref{fig:nonuni_training_data} shows the training data sampled from the manifold of the cusp.

\begin{figure}[ht]
    \centering
    \includegraphics[width=1\linewidth]{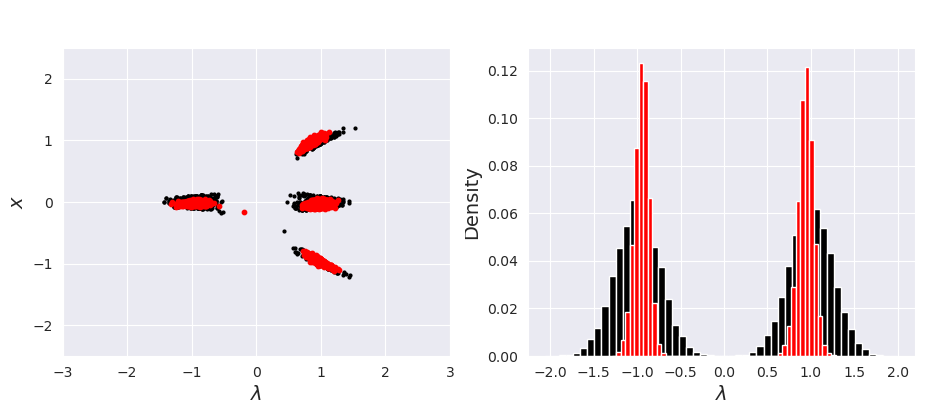}
    \caption{On the left, A slice of training data where $\mu \approx 0$ (black points) plotted with three-dimensional outputs from a cSGM (red points) where the conditioned values are $\mu = 0$. On the right, histograms of $\lambda$ values are shown in black (training data) and red (cSGM output).}
    \label{fig:nonuni_mcs_comp_with_hist}
\end{figure}

\begin{figure}[!htb]
    \centering
    \begin{subfigure}[b]{0.49\textwidth}
        \centering
        \includegraphics[width=\textwidth]{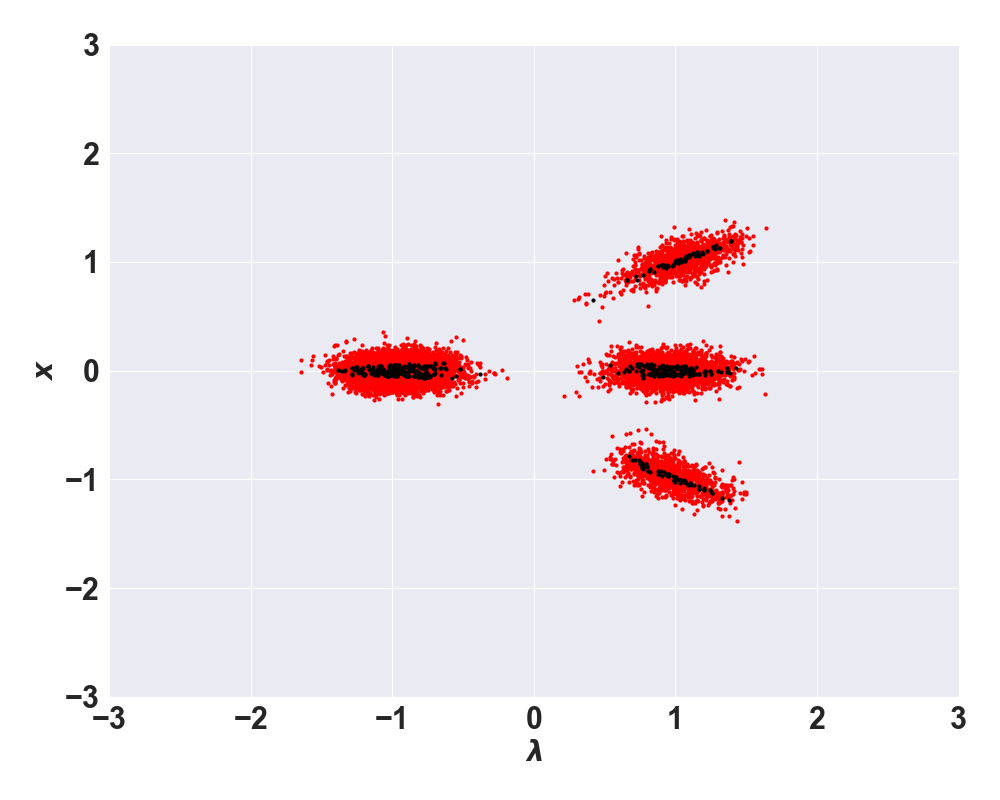}
        \caption{}
        \label{fig:suba1}
    \end{subfigure}
    \hfill
    \begin{subfigure}[b]{0.49\textwidth}
        \centering
        \includegraphics[width=\textwidth]{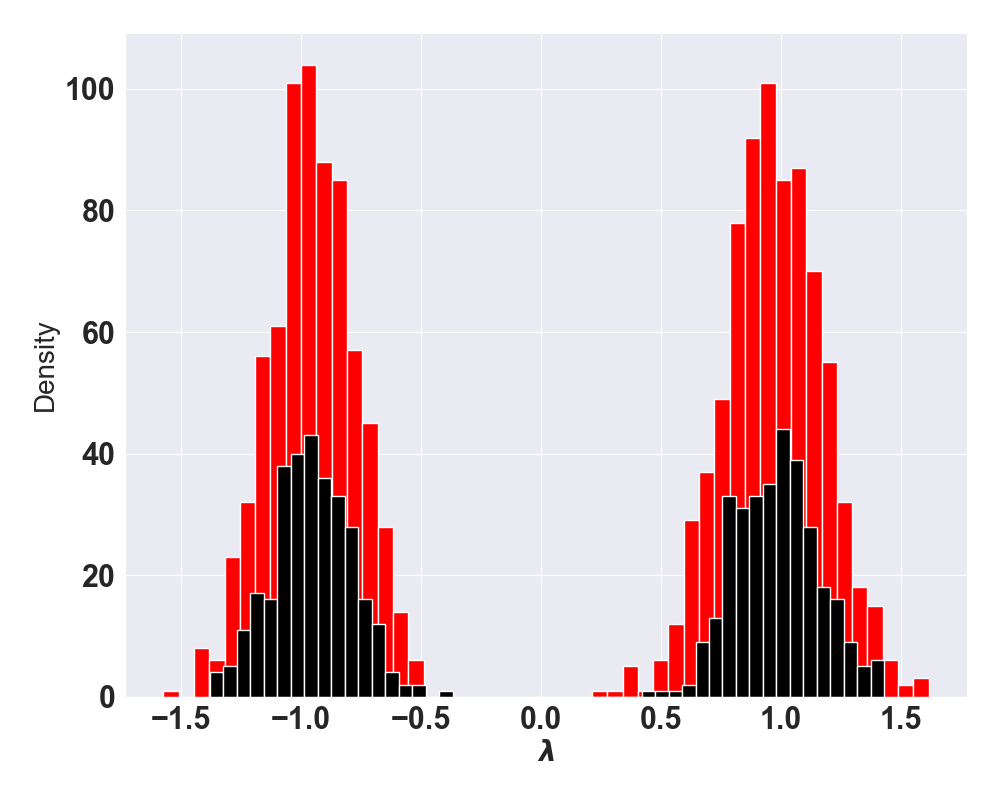}
        \caption{}
        \label{fig:suba2}
    \end{subfigure}
    \caption{(a) slice of training data where $\mu \approx 0$ (black points) plotted with three-dimensional outputs from the MCS-based SGM (red points) where the conditioned values are $\mu = 0$. On the right, histograms of $\lambda$ values are shown in black (training data) and red (cSGM output). In this case we see that the model performs better in learning the data distribution tales.}
    \label{fig:main}
\end{figure}

Following Algorithm \ref{alg:Training_SGM_alg}, a cSGM is trained on the new non-uniformly sampled manifold. Figure \ref{fig:nonuni_mcs_comp_with_hist} shows the output of the cSGM compared with the original training data for the conditional distribution where $\mu = 0$. The cSGM produces data at all possible steady states of the system, both pre- and post- bifurcation. The learned distribution in $\lambda$ is not an exact match, but it is not an unreasonable approximation. The results are depicted in Fig.~\ref{fig:main} for the MCS-based cSGM. In this case we see that the model performs better in learning the data distribution. Models and methods to improve the matching of the distributions for non-uniform cases are an avenue that will be explored in future work.

\section*{Conclusions}
\label{sec:conclusions}
\noindent
Conditional generative models hold significant parallels with physics-based sampling methods for systems of PDEs, such as initialization on slow manifolds \cite{init_on_man_VANDEKERCKHOVE, init_slow_man1}. We propose a framework for using generative models to initialize on reduced-order slow manifolds, and subsequently reconstruct full PDE profiles corresponding to the generated initial conditions. This framework provides value in high-dimensional systems where the manifold and its parameters are not readily apparent and when quickly discovering the low-dimensional behavior of a system is of high interest. cSGMs were demonstrated to be readily capable in producing data on the manifold that are conditioned on and consistent with one or more order parameters. Non-uniformly sampled manifolds were also explored, and cSGMs were demonstrated to be able to roughly approximate these nonstandard distributions on the manifold. Further investigation of methods to exactly match non-uniform distributions conditionally sampled on the manifold will are a direction for future work.


When comparing neural network-based conditional cSGMs with  MCS-based cSGMs, a key distinction lies in computational efficiency and flexibility. Neural network cSGMs leverage deep learning to learn complex conditional score functions from data, enabling expressive generative modeling across diverse conditions. However, training these models can be computationally intensive, and inference often requires iterative sampling schemes such as Langevin dynamics, which further increase runtime. In contrast, MCS-based cSGMs bypass the need for training by directly estimating the conditional distribution through repeated sampling from known or learned densities. As a result, they are significantly faster at runtime, especially when high-fidelity samples are needed in real time or in resource-constrained settings. While neural network cSGMs offer greater modeling flexibility and scalability to high-dimensional problems, MCS-based approaches provide a lightweight and efficient alternative when rapid conditional sampling is the priority.

\section*{Acknowledgements}
\label{sec:acknowledgements}

The research efforts of two authors (DG and IGK) were supported by the U.S. Department of Energy (DOE) under Grant No. DE-SC0024162.

\appendix
\section{SGM Training Details}

\noindent
As stated in Section \ref{sec:methods}, parameters and hyperparameters were chosen with the guidance of Karras et. al. \cite{karrasdiffparams}. For the Ornstein-Ulenbeck process, the SDE was chosen to have 1000 discretized time steps, and the parameters were chosen to be $\beta_{min} = 0.001$ and $\beta_d = 3$. The architecture of the neural network of the SGM is laid out in table \ref{tab:SDE_net}.
\begin{table}[H]
    \centering
    \caption{Architecture of the network for learning the drift of the SGM's SDE} 
    \begin{tabular}{!{\VRule[2pt]}c!{\VRule[2pt]}}
    \specialrule{2pt}{0pt}{0pt}
    input: data sample $x \in \mathbb{R}^N$, label $y \in \mathbb{R}^M$, time $t \in \mathbb{R}$ \\\hline
    concat(x,y,t) $\in \mathbb{R}^{N+M+1}$ \\\hline
    Dense(64);Tanh   \\\hline
    Dense(128);Tanh  \\\hline
    Dense(256);Tanh  \\\hline
    Dense(512);Tanh  \\\hline
    Dense(256);Tanh  \\\hline
    Dense(128);Tanh  \\\hline
    Dense(64);Tanh   \\\hline
    Dense($N$)  \\\specialrule{2pt}{0pt}{0pt}
    \end{tabular}
    \label{tab:SDE_net}
\end{table}

\bibliographystyle{unsrt}
\bibliography{bibliography}
\end{document}